\documentclass[11pt,a4paper]{article}
\usepackage[hyperref]{emnlp-ijcnlp-2019}
\usepackage{times}
\usepackage{latexsym}
\usepackage{amsmath}
\usepackage{url}
\usepackage{multirow}
\usepackage{graphicx}
\usepackage{graphics}
\usepackage{enumitem}
\usepackage{subcaption}
\usepackage{amssymb}
\usepackage[normalem]{ulem} 
\usepackage{makecell}

\usepackage{array}

\makeatletter
\newcommand{\thickhline}{%
    \noalign {\ifnum 0=`}\fi \hrule height 1pt
    \futurelet \reserved@a \@xhline
}

\title{Let's Ask Again: Refine Network for Automatic Question Generation}

\author{Preksha Nema$^{\dagger \ddagger}$\thanks{* The first two authors have contributed equally to this work.} \hspace{0.2cm} Akash Kumar Mohankumar$^{\dagger}$\footnotemark[1] \hspace{0.2cm} Mitesh M.  Khapra$^{\dagger \ddagger}$ \\
\bf{Balaji Vasan Srinivasan}$^{\bullet}$ \hspace{0.1cm} Balaraman Ravindran$^{\dagger \ddagger}$ \\
  $^\dagger$IIT Madras, India \hspace{0.1cm} 
  $^{\bullet}$Adobe Research\\
    $^{\ddagger}$ Robert Bosch Center for Data Science and Artificial Intelligence, IIT Madras \\
  {\tt \{preksha,miteshk,ravi\}@cse.iitm.ac.in} \\ {\tt makashkumar99@gmail.com} \hspace{0.1cm} {\tt \{balsrini\}@adobe.com } 
}
\date{}
\aclfinalcopy
\begin{document}

\maketitle
\begin{abstract}
In this work, we focus on the task of Automatic Question Generation (AQG) where given a passage and an answer the task is to generate the corresponding question. It is desired that the generated question should be (i) grammatically correct (ii) answerable from the passage and (iii) specific to the given answer. An analysis of existing AQG models shows that they produce questions which do not adhere to one or more of {the above-mentioned qualities}. In particular, the generated questions look like an incomplete draft of the desired question with a clear scope for refinement. {To alleviate this shortcoming}, we propose a method which tries to mimic the human process of generating questions by first creating an initial draft and then refining it. More specifically, we propose Refine Network (RefNet) which contains two decoders. The second decoder uses a dual attention network which pays attention to both (i) the original passage and (ii) the question (initial draft) generated by the first decoder. In effect, it refines the question generated by the first decoder, thereby making it more correct and complete. We evaluate RefNet on three datasets, \textit{viz.}, SQuAD, HOTPOT-QA, and DROP, and show that it outperforms existing state-of-the-art methods by 7-16\% on all of these datasets. Lastly, we show that we can improve the quality of the second decoder on specific metrics, such as, fluency and answerability by explicitly rewarding revisions that improve on the corresponding metric during training. The code has been made publicly available \footnote{https://github.com/PrekshaNema25/RefNet-QG}.


\end{abstract}

\section{Introduction}

\begin{table}[h]
    \small
    \begin{tabular}{p{7cm}}
        \Xhline{3\arrayrulewidth}
         \textbf{Passage 1:} Liberated by \textcolor{blue}{Napoleon's} army in 1806, Warsaw was made the capital of the newly created Duchy of Warsaw.\\
         \hline 
         \textbf{Generated Questions} \\
        \begin{tabular}{l|l}
                \textit{Baseline}& What was \textcolor{red}{the capital} of the newly \\ & duchy of Warsaw? \\
                \textit{RefNet}& \textcolor{magenta}{Who liberated} Warsaw in 1806?                  \\
                \textit{Reward-RefNet}& \textcolor{magenta}{Whose army} liberated Warsaw \\ &in 1806?          
        \end{tabular}\\
        
        \Xhline{3\arrayrulewidth}
         \textbf{Passage 2:} To fix carbon dioxide into sugar molecules in the process of photosynthesis, chloroplasts use \textcolor{blue}{an enzyme called rubisco} \\
         \hline 
         \textbf{Generated Questions} \\
            \begin{tabular}{l|l}
                \textit{Baseline}& What does chloroplasts use?\\
                \textit{RefNet}& What does chloroplasts use \textcolor{magenta}{to fix }\\&\textcolor{magenta}{carbon dioxide into sugar molecules?}\\
                \textit{Reward-RefNet}& What \textcolor{magenta}{do} chloroplasts use to fix  \\& carbon dioxide into sugar molecules?          
            \end{tabular}        \\
            
        \Xhline{3\arrayrulewidth}
    \end{tabular}
    \caption{Samples of generated questions from Baseline, RefNet and Reward-RefNet model on the SQuAD dataset.  Answers are shown in \textcolor{blue}{blue}}.
    \label{tab:example}
\end{table}

{Over the past few years, there has been a growing interest in Automatic Question Generation (AQG) from text - the task of generating a question from a passage and optionally an answer. AQG is used in curating Question Answering datasets,} enhancing user experience in conversational AI systems \cite{convagents} and {for creating educational materials \cite{good:question}}. {For the above applications}, it is {essential} that the questions are (i) grammatically correct (ii) answerable from the passage and (iii) specific to the answer. Existing approaches {focus on encoding the passage, the answer and the relationship between them using complex functions and then generate the question in \textit{one} single pass}. However, by carefully analysing the generated questions, we observe that these approaches tend to miss one or more of the {important aspects of the question}. For instance, in Table \ref{tab:example}, the question generated by the {single-pass} baseline model for the first passage is grammatically correct but is not specific to the answer. In the second example, the generated question is both syntactically incorrect and incomplete.
The above examples indicate that there is clear scope of improving the general quality of the questions. Additionally, the quality can be specifically improved in terms of aspects like: fluency (Example 2) and answerability (Example 1). One way to approach this is by re-visiting the passage and answer with the  aim to \textit{refine} the initial draft by  generating a better question in the second pass and then improving it with respect to a certain aspect.  We can draw a comparison between this process and how humans tend to write a rough initial draft first and then refine it over multiple passes, where the later revisions focus on improving the draft aiming at certain aspects like fluency or completeness. With this motivation, we propose \textbf{Refine Network (RefNet)}, which examines the initially generated question and performs a second pass to generate a \textit{revised} question. {Furthermore, we propose \textbf{Reward-RefNet} which uses explicit reward signals to achieve refinement focused on specific properties of the question such as fluency and answerability. }

Our RefNet is a seq2seq based model that comprises of two decoders: \textit{Preliminary} and \textit{Refinement} \textit{Decoder}. The Refinement Decoder takes the initial draft of the question generated by the Preliminary decoder as an input along with passage and answer, and generates {the refined} question by attending onto both the passage and the initial draft {using a Dual Attention Network}. The proposed dual attention aids RefNet to generate the final question by revisiting the appropriate parts of the input passage and initial draft. From Table \ref{tab:example}, we can infer that our RefNet model is able to generate better questions in the second pass by  fixing the errors in the initial draft. {Our} Reward-RefNet model uses REINFORCE {with a baseline algorithm to explicitly reward the Refinement Decoder for generating a better question as compared to the Preliminary Decoder based on certain desired parameters like fluency and answerability.} This leads to more answerable (see Reward-RefNet example for passage $1$ in Table \ref{tab:example}) and fluent (see Reward-RefNet example for passage $2$ in Table \ref{tab:example}) questions as compared to vanilla RefNet model. 

Our experiments show that the proposed RefNet model outperforms existing state-of-the-art models on the SQuAD dataset by $12.3$\% and $3.7$\% (on BLEU) given the relevant sentence and passage respectively. We also {achieve state-of-the-art results on HOTPOT-QA and DROP datasets with an improvement of $7.57$\% and $15.25$\% respectively over the single-decoder baseline (on BLEU)}. Our human evaluations further validate these results. We further analyze and explain the impact of including the Refinement Decoder by examining the interaction between both the decoders. Interestingly, we observe that the inclusion of the Refinement Decoder boosts the quality of the questions generated by the initial decoder also. Lastly, our human evaluation of the questions generated by Reward-RefNet corroborate empirical results, \textit{i.e.}, it improves the question \textit{w.r.t.} to fluency and answerability as compared to RefNet questions. 

\begin{figure*}
    \centering
    \includegraphics[width=0.9\textwidth,height=200pt]{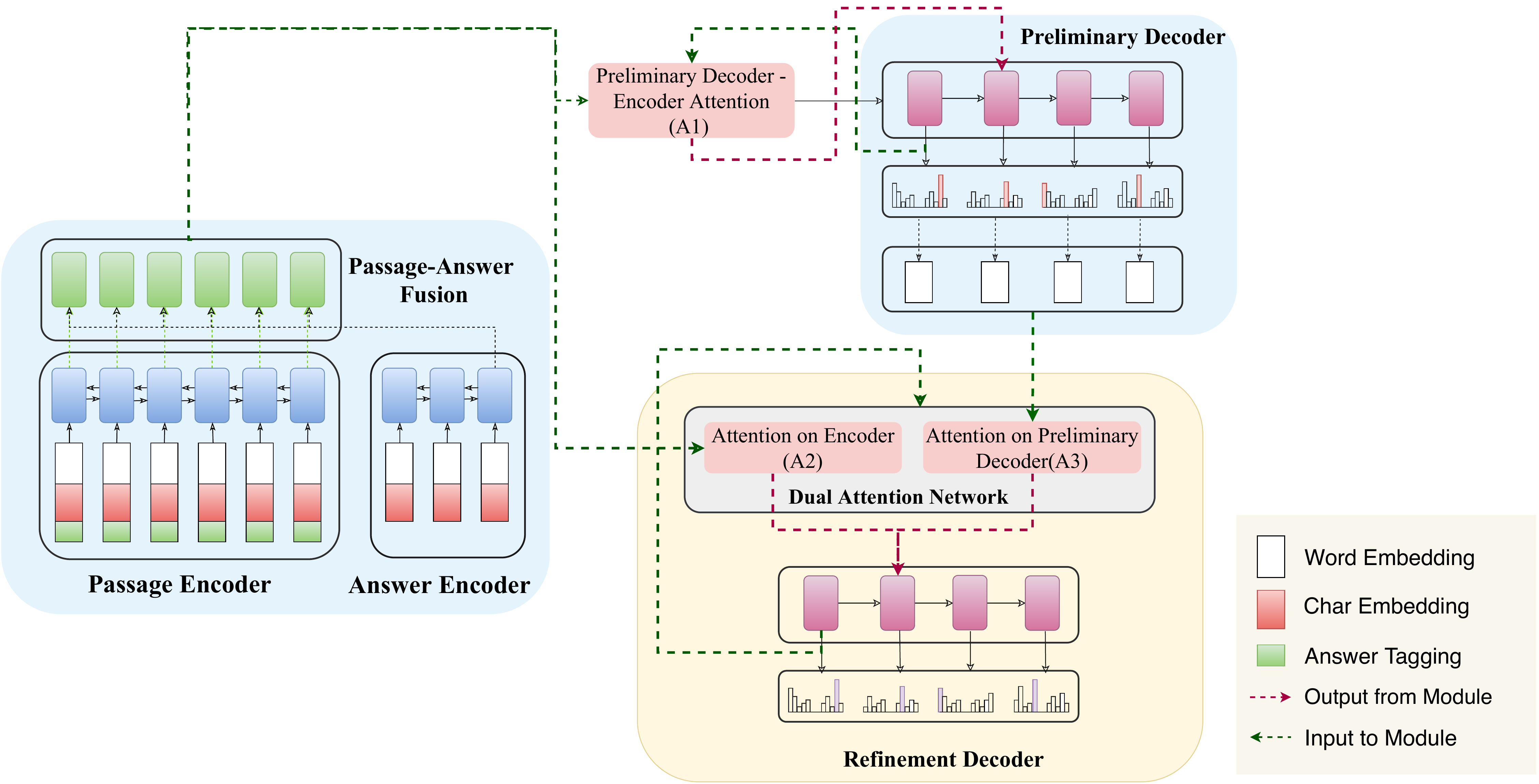}
    \caption{Our RefNet model with Preliminary and Refinement Decoder.}
    \label{fig:model}
\end{figure*}

\section{Refine Networks (RefNet) Model}
In this section, we discuss various components of our proposed model as shown in Figure \ref{fig:model}.
For a given passage $\mathbf{P} = \{w^p_1, \dots, w^p_m\}$  of length $m$ and answer $\mathbf{A} = \{w^a_1, \dots, w^a_n\}$ of length $n$, we first obtain \textit{answer-aware} latent representation, $\mathbf{U} = \{ \tilde{\mathbf{h}}^p_1,\dots,\tilde{\mathbf{h}}^p_m \}$, for every word of the passage and an answer representation $\mathbf{h}^a$ (as described in Section \ref{sec:nlu}).
We then generate an initial draft $\mathbf{\tilde{Q}} =$ 
$\{\tilde{q}_1, \dots, \tilde{q}_T\}$ by computing $\tilde{q}_t$ as
\begin{align}
      \nonumber \tilde{q}_t &=  \arg \max_{\tilde{q}} \prod_{t=1}^{l} \tilde{p}(\tilde{\textrm{q}}_t|\tilde{\textrm{q}}_{t-1}, \dots, \tilde{\textrm{q}}_1, \mathbf{U}, \mathbf{h}^a)
\end{align}

Here $\tilde{p}(.)$ is a probability distribution modeled using the Preliminary Decoder. We then refine the initial draft $\mathbf{\tilde{Q}}$ using the Refinement Decoder to obtain the refined draft $\mathbf{Q} = \{{q}_1, \dots {q}_T\}$:
\begin{align}
     \nonumber q_t &= \arg \max_q \prod_{t=1}^{l} p(\textrm{q}_t|\textrm{q}_{t-1},\dots,\textrm{q}_1, \mathbf{\tilde{Q}}, \mathbf{U}, \mathbf{h}^a)
\end{align}
We then use explicit rewards to enforce refinement on a desired metric, such as, fluency or answerability through our Reward-RefNet model. In the following sub-sections, we describe the passage encoder, preliminary and refinement decoders and our reward mechanism.


\subsection{Passage and Answer Encoder}
\label{sec:nlu}
We use a 3 layered encoder consisting of: (i) Embedding, (ii) Contextual and (iii) Passage-Answer Fusion layers as described below. To capture interaction between passage and answer, we ensure that the passage and answer representations are fused together at every layer.
{\flushleft \bf Embedding Layer:} In this layer, we compute a $d$-dimensional embedding for every word in the passage and the answer. This embedding is obtained by concatenating the word's Glove embedding \cite{glove} with its character based embedding as discussed in \cite{rcmodelbidaf}. Additionally, for passage words, we also compute a positional embedding based on the relative position of the word \textit{w.r.t.} the answer span as described in \cite{para}. For every passage word, this positional embedding is also concatenated to the word and character based embeddings. We discuss the impact of character embeddings and answer tagging in Appendix \ref{app:sec1}. 
In the subsequent sections, we will refer to embedding of the $i$-th passage word $w^p_i$ as $\mathbf{e(w^p_i)}$ and the $j$-th answer word $w^a_j$ as $\mathbf{e(w^a_j)}$.
{\flushleft \bf Contextual Layer:} In this layer, we compute a contextualized representation for every word in the passage by passing the word embeddings (as computed above) through a bidirectional-LSTM \cite{lstm}:
\begin{align}
     \nonumber \overrightarrow{
 \mathbf{h}^p_t} &= \text{LSTM}(\mathbf{e(w^p_t)}, \overrightarrow{\mathbf{h}^p}_{t-1})\; \;  \forall t \in [1,m]
\end{align}
where $\overrightarrow{
 \mathbf{h}^p_t}$ is the hidden state of the forward LSTM at time $t$.  We then concatenate the forward and backward hidden states as $\mathbf{h}^p_t =  [ \overrightarrow{
 \mathbf{h}^p_t}; \overleftarrow{
 \mathbf{h}^p_t} ]$.

The answer could correspond to a span in the passage. Let $j+1$ and $j+n$ be the start and end indices of the answer span in the passage respectively. We can thus refer to $\{\mathbf{h}^p_{j+1}, \dots, \mathbf{h}^p_{j+n}\}$ as the representation of the answer words in the context of the passage. We then obtain contextualized representations for the $n$ answer words by passing them through LSTM as follows: 

\begin{align}
    \nonumber \overrightarrow{\mathbf{h}^a_t} &= \text{LSTM}([\mathbf{e(w^a_t)}, \mathbf{h}^p_{j+t}],  \overrightarrow{\mathbf{h}^a}_{t-1}) \; \; \forall t \in [1,n] 
\end{align}

The final state $\mathbf{h}^a =  [ \overrightarrow{\mathbf{h}^a_n}; \overleftarrow{\mathbf{h}^a_n} ] $ of this Bi-LSTM is used as the answer representation in the subsequent stages. When the answer is not present in the passage, only $\mathbf{e(w^a_t)}$ is passed to the LSTM. 

{\flushleft \bf Passage-Answer Fusion Layer:} In this layer, we refine the representations of the passage words based on the answer representation  as follows: 
\begin{align}
    \nonumber \tilde{\mathbf{h}}^p_i = \tanh{(\mathbf{W}_u \left [ \, \mathbf{h}^p_i; \, \mathbf{h}^a; \, \mathbf{h}^p_i\odot \mathbf{h}^a \right ])} \; \; \forall i \in [1,m] 
\end{align}
Here $\mathbf{W}_u \in \mathbb{R}^{l \times 3l}$. $l$ is the hidden size of LSTM. This is similar to how \cite{rcmodelbidaf} capture interactions between passage and {question} for QA. We use $\mathbf{U} = \{ \tilde{\mathbf{h}}^p_1,\dots,\tilde{\mathbf{h}}^p_m \}$ as the {fused passage-answer} representation which is then used by our decoder(s) to generate the question $\mathbf{Q}$. 

\subsection{Preliminary and Refinement Decoders}
As discussed earlier, RefNet has  two decoders, \textit{viz}., Preliminary Decoder and Refinement Decoder, as described below: 

{\flushleft \bf Preliminary Decoder:} This decoder generates an initial draft of the question, one word at a time, using an LSTM as follows:
\begin{align}
    \nonumber \tilde{\mathbf{h}}^d_t &= \text{LSTM}([\mathbf{e_w(\tilde{q}_{t-1})};\tilde{\mathbf{c}}_{t-1};\mathbf{h}^a], \tilde{\mathbf{h}}^d_{t-1}) \\
    \tilde{\mathbf{c}}_t &= \sum_{i=1}^m \alpha^i_t \tilde{\mathbf{h}}_i^p
\end{align}
Here $\tilde{\mathbf{h}}^d_t$ is the hidden state at time $t$, $\mathbf{h}^a$ is the answer representation as computed above, $\tilde{\mathbf{c}}_{t-1}$ is an attention weighted sum of the contextualized passage word representations, $\alpha^i_t$ are parameterized and normalized attention weights \cite{Bahdanau}. {Let's} call this attention network as $\mathbf{A}_1$. 
$\mathbf{e_w(\tilde{q_t})}$ is the embedding of the word $\tilde{q_t}$. We obtain $\tilde{q_t}$ as:
\begin{align}
     \label{vocab}
     \tilde{q}_t = \text{arg}\max_{\tilde{q}} \; (\text{softmax}(\mathbf{W}_o[ \mathbf{W}_c[ \tilde{\mathbf{h}}^d_t;\tilde{\mathbf{c}}_t])),
\end{align}
where $\mathbf{W}_c$ is a $\mathbb{R}^{l \times 2l}$ matrix and $\mathbf{W}_o$ is the output matrix which projects the final representation to $\mathbb{R}^V$ where $V$ is the vocabulary size.

\textbf{Refinement Decoder}:  
Once the preliminary decoder generates the entire question, the refinement decoder uses it to generate an updated version of the question using a  \textbf{Dual Attention Network}. It first computes an attention weighted sum of the embeddings of the words generated by the first decoder as:
\begin{align*}
    \mathbf{g}_t = \sum_{i=1}^{T} \beta^t_i \mathbf{e_w(\tilde{q_i})}
\end{align*}
where $\beta^t_i$ are parameterized and normalized attention weights computed by attention network $\mathbf{A}_3$. Since the initial draft could be erroneous or incomplete, we obtain additional information from the passage instead of only relying on the output of the first decoder. We do so by computing a context vector $\mathbf{c}_t$ as
\begin{align*}
    \mathbf{c}_t = \sum_{i=1}^{m} \gamma^t_i \tilde{ \mathbf{h}}_i^p
\end{align*}
where $\gamma^t_i$ are parameterized and normalized attention weights computed by attention network $\mathbf{A}_3$. The hidden state of the refinement decoder at time $t$ is computed as follows:
\begin{align}
    \nonumber \mathbf{h}^d_t = \text{LSTM}([\mathbf{e(q_{t-1})};\mathbf{c}_{t-1};\mathbf{g}_{t-1};\mathbf{h^a}],\mathbf{h}^d_{t-1}) 
\end{align}
Finally, ${q}_t$ is predicted using
\begin{align}
    \nonumber {q}_t = \text{arg} \max_{q} (\text{softmax}( {\mathbf{W}}_o[ {\mathbf{W}'}_c[{\mathbf{h}}^d_t;{\mathbf{c}}_{t-1}; \mathbf{g}_{t-1} ]))
\end{align}
where $\mathbf{W}'_c$ is a weight matrix and $\mathbf{W}_o$ is the output matrix which is shared with the Preliminary decoder (Equation \ref{vocab}). Note that RefNet generates two variants of the question : initial draft  $\tilde{ \mathbf{Q}}$ and final draft $\mathbf{{Q}}$. We compare these two versions of the generated questions in Section \ref{sec:results}.

\subsection{Reward-RefNet}
\label{sec:reward}
Next, we address the following question: \textit{Can the refinement decoder be explicitly rewarded for generating a question which is better than that generated by the preliminary decoder on certain desired parameters?} For example, \cite{qbleu} define fluency and answerability as desired qualities in the generated question. They evaluate fluency using BLEU score and answerability using a score which captures whether the question contains the required \{named entities, {important words, function words, question types}\} (and is thus answerable). We use these fluency and answerability scores proposed by \cite{qbleu} as reward signals. We first compute the reward $r(\mathbf{\tilde{Q}})$ and  $r(\mathbf{Q})$ for the question generated by the preliminary and refinement decoder respectively. We then use ``REINFORCE with a baseline" algorithm \cite{Williams} to reward Refinement Decoder using the Preliminary Decoder's reward  $r(\mathbf{\tilde{Q}})$ as the baseline. More specifically, given the Preliminary Decoder's generated word sequence $\mathbf{\tilde{Q}} = \{\tilde{q}_1, \tilde{q}_2,\dots,\tilde{q}_T\}$ and the Refinement Decoder's generated word sequence $\mathbf{Q} = \{q_1, q_2,\dots,q_T\}$ obtained from the distribution $p(\textrm{q}_t|\textrm{q}_{t-1},\dots,\textrm{q}_1, \mathbf{\tilde{Q}}, \mathbf{U}, \mathbf{h}^a)$, the training loss is defined as follows
\begin{align}
 \nonumber L(\mathbf{Q}) =& (r(\mathbf{Q}) - r(\mathbf{\tilde{Q}})) \cdot \\
 \nonumber & \sum_{t=1}^{T} \textrm{log} \; p({q}_t|\ {q}_{t-1},\dots, {q}_1, \mathbf{\tilde{Q}}, \mathbf{U}, \mathbf{h}^a)
 \end{align}
where $r(\mathbf{Q})$ and $r(\mathbf{\tilde{Q}})$  \textcolor{blue}are the rewards obtained by comparing with the reference question $\mathbf{Q^*}$. As mentioned, this reward $r(.)$ can be the fluency score or answerability score as defined by \cite{qbleu}.


\subsection{Copy Module}
Along with the above-mentioned three modules, we adopt the pointer-network and coverage mechanism from \cite{getothepoint}. We use it to (i) handle Out-of-Vocabulary words and (ii) avoid repeating phrases in the generated questions.

\begin{table*}[]
\resizebox{\textwidth}{!}{
\begin{tabular}{|l|lllllll|l|}
\Xhline{3\arrayrulewidth}
\multirow{2}{*}{\textbf{Dataset}}                                                          & \multicolumn{1}{l|}{\multirow{2}{*}{\textbf{Model}}} & \multicolumn{6}{c|}{\textbf{n-gram}}                                                                                                                                                                         &  \\ \cline{3-8} 
                                                                                           & \multicolumn{1}{l|}{}                                & \multicolumn{1}{l|}{BLEU-1}         & \multicolumn{1}{l|}{BLEU-2}         & \multicolumn{1}{l|}{BLEU-3}         & \multicolumn{1}{l|}{BLEU-4}         & \multicolumn{1}{l|}{ROUGE-L}        & METEOR         &  QBLEU4               \\ \Xhline{3\arrayrulewidth}
\multirow{5}{*}{\textbf{\begin{tabular}[c]{@{}l@{}}SQuAD\\ (Sentence Level)\end{tabular}}} 
  &                           \multicolumn{1}{l|}{\cite{answerfocuesdqg}}                          & \multicolumn{1}{l}{$43.02$}          & \multicolumn{1}{l}{$28.14$}          & \multicolumn{1}{l}{$20.51$}          & \multicolumn{1}{l}{$15.64$}          & \multicolumn{1}{l}{-}          & -         & -                  \\ 
   &                           \multicolumn{1}{l|}{\cite{para}}                         & \multicolumn{1}{l}{$44.51$}          & \multicolumn{1}{l}{$29.07$}          & \multicolumn{1}{l}{$21.06$}          & \multicolumn{1}{l}{$15.82$}          & \multicolumn{1}{l}{$44.24$}          & $19.67$          & -                 \\ 
   &                           \multicolumn{1}{l|}{\cite{Kim2019ImprovingNQ}}                          & \multicolumn{1}{l}{-}          & \multicolumn{1}{l}{-}          & \multicolumn{1}{l}{-}          & \multicolumn{1}{l}{$16.17$}          & \multicolumn{1}{l}{-}          & -         & -                  \\ \cline{2-9}

                                                                & \multicolumn{1}{l|}{EAD}                             & \multicolumn{1}{l}{$44.74$}          & \multicolumn{1}{l}{$29.79$}          & \multicolumn{1}{l}{$22.00$}          & \multicolumn{1}{l}{$16.84$}          & \multicolumn{1}{l}{$44.78$}          & $20.60$          &$ 24.7$                  \\
                                                                                         
                                                                 &                           \multicolumn{1}{l|}{RefNet}                          & \multicolumn{1}{l}{$\mathbf{47.27}$}         & \multicolumn{1}{l}{$\mathbf{31.88}$}          & \multicolumn{1}{l}{$\mathbf{23.65}$}          & \multicolumn{1}{l}{$\mathbf{18.16}$}          & \multicolumn{1}{l}{$\mathbf{47.14}$}          &  $\mathbf{23.40}$          & $\mathbf{27.4}$                  \\ 
                                                                                  \Xhline{3\arrayrulewidth}

\multirow{3}{*}{\textbf{\begin{tabular}[c]{@{}l@{}}SQuAD\\ (Passage Level)\end{tabular}}}  
                                                     &                           \multicolumn{1}{l|}{\cite{para}}                          & \multicolumn{1}{l}{$45.07$}          & \multicolumn{1}{l}{$29.58$}          & \multicolumn{1}{l}{$21.60$}          & \multicolumn{1}{l}{$16.38$}          & \multicolumn{1}{l}{$44.48$}          & $20.25$          & -                  \\ \cline{2-9}

& \multicolumn{1}{l|}{EAD}                             & \multicolumn{1}{l}{$44.61$}          & \multicolumn{1}{l}{$29.37$}          & \multicolumn{1}{l}{$21.50$}          & \multicolumn{1}{l}{$16.36$}          & \multicolumn{1}{l}{$43.95$}          & $20.11$         &$ 24.2$                  \\
     & \multicolumn{1}{l|}{RefNet}                          & \multicolumn{1}{l}{$\mathbf{46.41}$}          & \multicolumn{1}{l}{$\mathbf{30.66}$}          & \multicolumn{1}{l}{$\mathbf{22.42}$}          & \multicolumn{1}{l}{$\mathbf{16.99}$}          & \multicolumn{1}{l}{$\mathbf{45.03}$}          & $\mathbf{21.10}$          & $\mathbf{26.6}$                  \\

                                                            
                                                                  \Xhline{3\arrayrulewidth}

\multirow{3}{*}{\textbf{HOTPOT}} &                           \multicolumn{1}{l|}{\cite{para}*}                          & \multicolumn{1}{l}{$45.29$}          & \multicolumn{1}{l}{$32.06$}          & \multicolumn{1}{l}{$24.43$}          & \multicolumn{1}{l}{$19.29$}          & \multicolumn{1}{l}{$40.40$}          & $19.29$         &$ 25.7$              \\ \cline{2-9}

& \multicolumn{1}{l|}{EAD}                                                            & $\mathbf{46.00}$         & $32.47$          & $24.82$          & $19.68$          & $41.52$          & $23.27$          &    $ 26.2$               \\
                                       & \multicolumn{1}{l|}{RefNet}                                                          & $45.45$          & $\mathbf{33.13}$          & $\mathbf{26.05}$          & $\mathbf{21.17}$          & $\mathbf{43.12}$          & $\mathbf{25.81}$          & $\mathbf{28.7}$  \\
                                       \Xhline{3\arrayrulewidth}
\multirow{3}{*}{\textbf{DROP Dataset}} &                           \multicolumn{1}{l|}{\cite{para}*}                          & \multicolumn{1}{l}{$39.56$}          & \multicolumn{1}{l}{$29.19$}          & \multicolumn{1}{l}{$22.53$}          & \multicolumn{1}{l}{$18.07$}          & \multicolumn{1}{l}{$45.01$}          & $19.68$          & $ 31.4$                 \\ \cline{2-9}

&\multicolumn{1}{l|}{EAD}                                                            & $39.21$          & $29.10$          & $22.65$          & $18.42$          & $45.07$          & $19.56$          &$ 31.8$                 \\
                                       & \multicolumn{1}{l|}{RefNet}                                                    & $\mathbf{42.81}$         & $\mathbf{32.63}$        & $\mathbf{25.78}$        & $\mathbf{21.23}$       & $\mathbf{47.49}$       & $\mathbf{22.25}$          &   $\mathbf{33.6}$                \\
                                       
\Xhline{3\arrayrulewidth}
\end{tabular}}
\caption{Comparsion of RefNet model with existing approaches and EAD model. Here * denotes our implementation of the corresponding work.}
\label{tab:results1}
\end{table*}

\section{Experimental Details}
In this section, we discuss (i) the datasets for which we tested our proposed model, (ii) implementation details and (iii) evaluation metrics used to compare our model with the baseline and existing works.

\subsection{Datasets}
\textbf{SQuAD} \cite{datasquad}: It contains $100$K (question, answer) pairs obtained from $536$ Wikipedia articles, where the answers are a span in the passage. For SQuAD, AQG has been tried from both sentences and passages. In the former case, only the sentence which contains the answer span is used as input, whereas in the latter case the entire passage is used. We use the same train-validation-test splits as used \textcolor{blue}{in} \cite{para}. 

\noindent\textbf{Hotpot QA} \cite{yang2018hotpotqa} : Hotpot-QA is a multi-document and multi-hop QA dataset. Along with the triplet (P, A, Q), the authors also provide supporting facts that potentially lead to the answer. The answers here are either yes/no or answer span in P. We concatenate these supporting facts to form the passage. We use $10$\% of the training data for validation and use the original dev set as test set.

\noindent\textbf{DROP} \cite{Dua2019DROP}: The DROP dataset is a reading comprehension benchmark which requires discrete reasoning over passage. It contains $96$K questions which require discrete operations such as addition, counting, or sorting to obtain the answer. We use $10$\% of the original training data for validation and use the original dev set as test set.

\subsection{Implementation Details}
We use $300$ dimensional pre-trained Glove word embeddings, which are fixed during training. For character-level embeddings, we initially use a $20$ dimensional embedding for the characters which is then projected to $100$ dimensions. For answer-tagging, we use embedding size of $3$. The hidden size for all the LSTMs is fixed to $512$. We use 2-layer, 1-layer and 2-layer stacked BiLSTM for the passage encoder, answer encoder and the decoders (both) respectively. We take the top $30,000$ frequent words as the vocabulary. We use Adam optimizer with a learning rate of $0.0004$ and train our models for $10$ epochs using cross entropy loss. For the Reward-RefNet model, we fine-tune the pre-trained model with the loss function mentioned in Section \ref{sec:reward} 
for $3$ epochs. The best model is chosen based on the BLEU \cite{bleu} score on the validation split. For all the results we use beam search decoding with a beam size of $5$.

\subsection{Evaluation}
 We evaluate our models, based on $n$-gram similarity metrics BLEU \cite{bleu}, ROUGE-L \cite{rouge}, and METEOR \cite{meteor} using the package released in \cite{bleupackage}\footnote{\url{https://github.com/Maluuba/nlg-eval}}. We also quantify the answerability of our models using QBLEU-4\footnote{\url{https://github.com/PrekshaNema25/Answerability-Metric}}\cite{qbleu}.\\

\section{Results and Discussions}
\label{sec:results}

\label{sec:results}

In this section, we present the results and analysis of our proposed model RefNet.  Throughout this section, we refer to our models as follows:
{\flushleft \bf Encode-Attend-Decode (EAD) model} is our single decoder model containing the encoder and the Preliminary Decoder described earlier. Note that the performance of this model is comparable to our implementation of the model proposed in \cite{para}.
{\flushleft \bf Refine Network (RefNet) model} includes the encoder, the Preliminary Decoder and the Refinement Decoder.\\
We will (i) compare RefNet's performance with EAD and existing models across all the mentioned datasets (ii) report human evaluations to compare RefNet and EAD (iii) analyze Refinement and Preliminary Decoders iv) present the performance of Reward RefNet with two different reward signal (fluency and answerability).

\subsection{RefNet's performance across datasets}
    In Table \ref{tab:results1}, we compare the performance of RefNet with existing single decoder architectures across different datasets. On BLEU-4 metric, RefNet beats the existing state-of-the-art model by $12.30$\%, $9.74$\%, $17.48$\%, and $3.71$\% respectively on SQuAD (sentence), HOTPOT-QA, DROP and SQuAD (passage) dataset. Also it outperforms EAD by $7.83$\%, $7.57$\%, $15.25$\% and $3.85$\% respectively on SQuAD (sentence), HOTPOT-QA, DROP and SQuAD (passage). 
    In general, RefNet is consistently better than existing models across all $n$-gram scores (BLEU, ROUGE-L and METEOR). Along with $n$-gram scores, we also observe improvements on Q-BLEU4 as well, which as described earlier, gives a measure of both answerability and fluency. 



\subsection{Human Evaluations}

We conducted human evaluations to analyze the quality of the questions produced by EAD and RefNet. We randomly sampled $500$ questions generated from the SQuAD (sentence level) dataset and asked the annotators to compare the quality of the generated questions. The annotators were shown a pair of questions, one generated by EAD and one by RefNet from the same sentence, and were asked to decide which one was better in terms of Fluency, Completeness, and Answerability. They were allowed to skip the question pairs where they could not make a clear choice. Three annotators rated each question and the final label was calculated based on majority voting. We observed that the RefNet model outperforms the EAD model across all three metrics. Over $68.6$\%, $66.7$\% and $64.2$\% of the generated questions from RefNet were respectively more fluent, complete and answerable when compared to the EAD model. However, there are some cases where EAD does better than RefNet. For example, in Table \ref{tab:basebetter}, we show that while trying to generate a more elaborate question, RefNet introduces an additional phrase \textit{``in the united''} which is not required. Due to such instances, annotators preferred the EAD model in around $30$\% of the instances.
\begin{table}
\resizebox{\columnwidth}{!}{
    \small
    \begin{tabular}{p{7cm}}
        \hline 
         \textbf{Passage:} Before the freeze ended in 1952, there were only 108 existing television stations in the United States; a few major cities (such as Boston) had only \textcolor{blue}{two} television stations, ...\\
         \hline 
         \textbf{Questions} \\
          \textit{EAD}: how many television stations existed in boston ?\\
          \textit{RefNet}:   how many television stations did boston have in the united ?\\
         \hline
    \end{tabular}}
    \caption{An example where EAD model was better than RefNet. The ground truth answers are shown in \textcolor{blue}{blue}.}
    \label{tab:basebetter}
\end{table}

\begin{table}[]
\resizebox{\columnwidth}{!}{
\begin{tabular}{|l|l|l|l|}
\hline
Model                         & \textbf{Decoder}       & \textbf{BLEU-4} & \textbf{QBLEU-4}  \\ \hline
\multirow{2}{*}{without $\mathbf{A_3}$}  & \textit{RefNet}    & $17.16$           & $25.80$          \\ \cline{2-4} 
                              & \textit{Initial Draft} & $17.59$           & $26.00$                      \\ \hline
\multirow{2}{*}{with $\mathbf{A_3}$} & \textit{RefNet}    & $18.37$           & $27.40$                     \\ \cline{2-4} 
                              & \textit{Initial Draft} & $17.89$           & $26.00$                       \\ \hline
\end{tabular}
}
\caption{Comparison between Preliminary Decoder and Refinement Decoder in RefNet Model for SQuAD Sentence Level QG.}
\label{tab:d1d2}
\end{table}

\begin{table*}
\resizebox{\textwidth}{!}{
\small
\begin{tabular}{|l|l|l}
\hline
\multicolumn{1}{l}{\textbf{Sample: }}  & \multicolumn{1}{l}{\begin{tabular}[c]{@{}l@{}}\textbf{Sentence:} For instance , the language \{ xx | x is any binary string \} can be solved in \\ \textcolor{blue}{linear time} on a multi-tape Turing machine ,  but necessarily requires \\ quadratic time in the model of single-tape Turing machines .\\ \textbf{Reference Question:} A multi-tape Turing machine requires what type of time for a solution ?\end{tabular}}  \\ \hline
\multicolumn{1}{l}{\textbf{with $\mathbf{A}_3$}} & \multicolumn{1}{l}{\begin{tabular}[c]{@{}l@{}}\textbf{Refinement Decoder}: in what time can the language be solved on a multi-tape turing machine ?\\ \textbf{Preliminary Decoder:} in what time can the language be solved ?\end{tabular}}                                                                                                                                                                             \\ \hline
 \multicolumn{1}{l}{\textbf{without $\mathbf{A}_3$}} & \multicolumn{1}{l}{\begin{tabular}[c]{@{}l@{}}\textbf{Refinement Decoder:} in what time can the language \{ xx | x x x x is any binary string ?\\ \textbf{Preliminary Decoder:} in what time can the language | x x x x is solved ?\end{tabular}}                                                                                                                                                                             \\ \hline

\end{tabular}
}
\caption{Generated samples by Preliminary Decoder and Refinement Decoder in RefNet model.}
\label{tab:generatedsamples1}
\end{table*}

\subsection{Analysis of Refinement Decoder and Preliminary Decoder}
The two decoders impact each other through two paths: (i) indirect path, where they share the encoder and the output projection to the vocabulary $V$, (ii) direct path, via the dual attention network, where the initial draft of the question is attended by the Refinement Decoder. 
When RefNet has only indirect path,  we can infer from row 1 of Table \ref{tab:d1d2} that the performance of Preliminary Decoder improves when compared to the EAD model ($16.84$ v/s $17.59$ BLEU). This suggests that generating two variants of the question improves the performance of the first decoder pass as well. This is perhaps due to the additional feedback that the shared encoder and output layer get from the Refinement Decoder. When we add the direct path (attention network) between the two decoders, the performance of the Refinement Decoder improves as compared to the Preliminary Decoder as shown in rows 3 and 4 of the Table \ref{tab:d1d2}\\
\noindent \textbf{Comparison on Answerability:} We also evaluate both the initial and refined draft using QBLEU4. As discussed earlier, Q-Metric measures Answerability using four components, \textit{viz.}, Named Entities, Important Words, Function Words, and Question Type. We observe that the increase in Q-Metric for refined questions is because the RefNet model can correct/add the relevant Named Entities in the question. In particular, we observe that the Named Entity component score in Q-Metric increases from $32.42$ for the first draft to $37.81$ for the refined draft. \\
\begin{figure}
    \centering
    \includegraphics[scale=0.45]{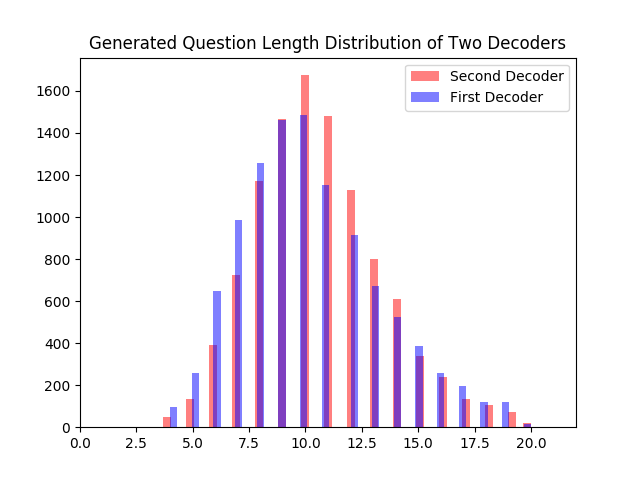}
    \caption{Generated Question Length Distribution for Preliminary Decoder (First Decoder) and Refinement Decoder (Second Decoder).}
    \label{fig:lengthdist}
\end{figure}
\noindent \textbf{Qualitative Analysis:} Figure~\ref{fig:lengthdist} shows that the RefNet model indeed generates more elaborate questions when compared to the Preliminary Decoder. As shown in Table~\ref{tab:generatedsamples1}, the quality of the refined question is better than the initial draft of the questions. Here RefNet adds the phrase ``multi-tape Turing Machine," {(row 2)} which removes any ambiguity in the question.

\subsection{Analysis of Reward-RefNet}
In this section, we analyze the impact of employing different reward signals in Reward-RefNet. As discussed earlier in section \ref{sec:reward}, we use fluency and answerability scores as reward signals. As shown in Table \ref{tab:rewardsignal}, when BLEU-4 (fluency) is used as a reward signal, there is improvement in BLEU-4 scores of Reward-RefNet as compared to RefNet model. We validated these results through human evaluations across $200$ samples. Annotators prefer the Reward-RefNet model in $67$\% of the cases for fluency. Similarly when we use Answerability score as a reward signal, answerability improves for the model and annotators prefer the Reward-RefNet in $70$\% of the cases for answerability. The performance of Reward-RefNet on fluency and answerability is similar for other datasets (see Appendix \ref{app:sec2}).

\begin{table}
\resizebox{\columnwidth}{!}{

\begin{tabular}{l|ll|l|l}
\hline
\multicolumn{1}{c|}{\multirow{2}{*}{\textbf{Model}}} & \multicolumn{2}{c|}{\textbf{\begin{tabular}[c]{@{}c@{}}BLEU \\ Reward Signal\end{tabular}}} & \multicolumn{2}{c}{\textbf{\begin{tabular}[c]{@{}c@{}}Answerability\\ Reward Signal\end{tabular}}} \\ \cline{2-5} 
\multicolumn{1}{c|}{}                                & \multicolumn{1}{p{1.5cm}|}{BLEU4}    & \begin{tabular}[c]{@{}l@{}}\%prefer-\\ ence\end{tabular}   & Ans.                   & \begin{tabular}[c]{@{}l@{}}\%prefer\\ ence\end{tabular}                   \\ \hline
RefNet                                               & \multicolumn{1}{l|}{$18.37$}    & $32.9$\%                                                          & $36.9$                   & $30$\%                                                                         \\ \hline
Reward-RefNet                                        & \multicolumn{1}{l|}{$18.52$}    &    $67.1$\%                                                        & $37.5$                   & $70$\%                                                                          \\ \hline
\end{tabular}}
\caption{Impact of Reward-RefNet on fluency and answerability. \%preference denotes the percentage of times annotators prefer the generated output from the model for fluency in case of BLEU Reward signal and answerability in case of Answerability Reward signal.}
\label{tab:rewardsignal}
\end{table}

\begin{table}
\resizebox{\columnwidth}{!}{
    \small
    \begin{tabular}{p{7cm}}
        \hline 
         \textbf{Passage:} \textcolor{blue}{Cost engineers and estimators} apply expertise to relate the work and materials
         involved to a proper valuation \\
         \hline 
         \textbf{Questions} \\
          \textit{Generated}: Who apply expertise to relate the work and materials involved to a proper valuation ? \\
          \textit{True}: Who applies expertise to relate the work and materials involved to a proper valuation ?\\
         \hline
    \end{tabular}}
    \caption{An example of question with significant overlap with the passage. The answer is shown in \textcolor{blue}{blue}.}
    \label{tab:originality}
\end{table}

\begin{figure}
    \centering
    \includegraphics[scale=0.45]{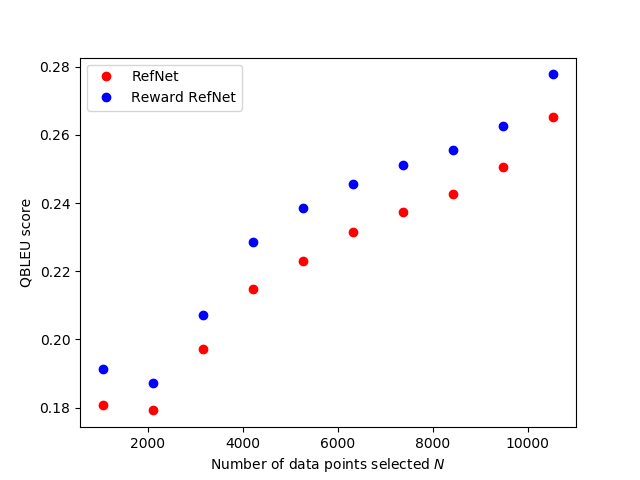}
    \caption{Originality Analysis: Plot of Q-BLEU score vs $N$ - the number points selected.}
    \label{fig:qbleu_originality}
\end{figure}

\noindent\textbf {Case Study: Originality of the Questions}\\
We observe that current state-of-the-art models perform very well in terms of BLEU/QBLEU scores when the actual question has significant overlap with the passage. For example, consider a passage from the SQuAD dataset in Table \ref{tab:originality}, where except the question word \textit{who}, the model sequentially copies everything from the passage and achieves a QBLEU score of $92.4$. However, the model performs poorly in situations where the true question is novel and does not contain a large sequence of words from the passage itself. In order to quantify this, we first sort the true questions based on its BLEU-2 overlap with the passage in ascending order. We then select the first $N$ true questions and compute the QBLEU score with the generated questions. The results are shown in red in Figure \ref{fig:qbleu_originality}. Towards the left, where there are true questions with low overlap with the passage, the performance is poor, but it gradually improves as the overlap increases.

The task of generating questions with high \textit{originality} (where the model phrases the question in its own words) is a challenging aspect of AQG since it requires complete understating of the semantics and syntax of the language. In order to improve questions generated on \textit{originality}, we explicitly reward our model for having low $n$-gram score with the passage as compared to the initial draft. As a result we observe that with Reward-RefNet(Originality), there is an improvement in the performance where the overlap with the passage was less (as shown in blue in Figure \ref{fig:qbleu_originality}).As shown in Table \ref{tab:originality_better}, although both questions are answerable given the passage, the question generated from Reward-RefNet(Originality) is better.

\begin{table}
\resizebox{\columnwidth}{!}{
    \small
    \begin{tabular}{p{7cm}}
        \hline 
         \textbf{Passage:} McLetchie was elected on the Lothian regional list and the Conservatives suffered a net loss of five seats , with leader \textcolor{blue}{Annabel Goldie} \textcolor{red}{claiming that their support had held firm}, nevertheless, she too announced she would step down as leader of the party. \\
         \hline 
         \textbf{Questions} \\
         \textit{True}: Who announced she would step down as leader of the    Conservatives ? \\
          \textit{RefNet}: who \textcolor{red}{claiming that their support had held firm ?} \\
          \textit{Reward-RefNet}: who was the leader of the conservatives? \\
         \hline
    \end{tabular}}
    \caption{An example where Reward-RefNet(Originality) is better than RefNet.}
    \label{tab:originality_better}
\end{table}

\section{Related Work}
Early works on Question Generation were essentially rule based systems \cite{good:question, DBLP:conf/aied/MostowC09, DBLP:conf/enlg/LindbergPNW13, DBLP:conf/acl/LabutovBV15}. Current models for AQG are based on the encode-attend-decode paradigm and they either generate questions from the passage alone  \cite{modelwheretofocus,learningtoask,Yao2018TeachingMT} or from the passage and a given answer (in which case the generated question must result in the given answer). Over the past couple of years, several variants of the encode-attend-decode model have been proposed. For example, 
\cite{Zhou2018SequentialCN} proposed a sequential copying mechanism to explicitly select a sub-span from the passage. Similarly, \cite{para} mainly focuses on efficiently incorporating paragraph level content by using Gated Self Attention and Maxout pointer networks. Some works \cite{modeltexttotext} even use Question Answering as a metric to evaluate the generated questions. There has also been some work on generating questions from images \cite{qgvae, qgvqa} and from knowledge bases \cite{sarathfactoid, datakbgenerationeacl}. The idea of multi pass decoding which is central to our work has been used by \cite{delibnet} for machine translation and text summarization albeit with a different objective. 
Some works have also augmented seq2seq models \cite{Rennie2017SelfCriticalST,Paulus2018ADR,Song2017AUQ} with external reward signals using REINFORCE with baseline algorithm \cite{Williams}. The typical  rewards used in these works are BLEU and ROUGE scores. Our REINFORCE loss is different from the previous ones as it uses the first decoder's reward as the baseline instead of reward of the greedy policy.

\section{Conclusion and Future Work}
In this work, we proposed \textbf{Refine Networks (RefNet)} for Question Generation to focus on refining and improving the initial version of the generated question. Our proposed RefNet model consisting of a Preliminary Decoder and a Refinement Decoder with Dual Attention Network outperforms the existing state-of-the-art models on the SQuAD, HOTPOT-QA and DROP datasets. Along with automated evaluations, we also conducted human evaluations to validate our findings. We further showed that using Reward-RefNet improves the initial draft on specific aspects like fluency, answerability and originality.  As a future work, we would like to extend RefNet to have the ability to decide whether a refinement is needed on the generated initial draft. 

\section*{Acknowledgements}
We thank Amazon Web Services for providing free GPU compute and Google for supporting Preksha Nema's contribution in this work through Google Ph.D. Fellowship programme. We would like to acknowledge Department of Computer Science and Engineering, IIT Madras and Robert Bosch Center  for  Data  Sciences  and  Artificial  Intelligence, IIT Madras (RBC-DSAI) for providing us sufficient resources. We would also like to thank Patanjali SLPSK, Sahana Ramnath, Rahul Ramesh, Anirban Laha, Nikita Moghe and the anonymous reviewers for their valuable and constructive suggestions. 


\bibliography{emnlp-ijcnlp-2019}
\bibliographystyle{acl_natbib.bst}
\appendix
\section{Impact of Various Embeddings}
\label{app:sec1}
We perform an ablation study to identify the impact of various word embeddings used in RefNet. When character embedding is not used in RefNet, the performance on SQuAD sentence-level drops from $18.16$ to $17.97$ BLEU-4 score. Meanwhile, when positional embeddings are dropped the performance decreases to $17.87$ BLEU-4 score.

\section{Reward-RefNet on Various Datasets}
\label{app:sec2}
Table \ref{tab:app} shows the comparison between RefNet and Reward-RefNet on BLEU-4 score and answerability score when the respective scores are used as rewards in Reward-RefNet. We can infer from Table \ref{tab:app} that there is improvement in fluency and answerability across all the datasets.
 
\begin{table}[h]
\resizebox{\columnwidth}{!}{
\begin{tabular}{|l|l|l|l|}
\hline
\textbf{Datasets}                                                                & \multicolumn{1}{c|}{\textbf{Model}} & \multicolumn{1}{c|}{\textbf{\begin{tabular}[c]{@{}c@{}}BLEU-4\\ Reward Signal\end{tabular}}} & \textbf{\begin{tabular}[c]{@{}l@{}}Answerability \\ Reward Signal\end{tabular}} \\ \hline
\multirow{2}{*}{\begin{tabular}[c]{@{}l@{}}\textbf{SQuAD}\\ \textbf{(Passage Level)}\end{tabular}} & RefNet                              & 16.99                                                                                        & 26.6                                                                            \\ \cline{2-4} 
                                                                                 & Reward-RefNet                       & 17.11                                                                                        & 27.3                                                                            \\ \hline
\multirow{2}{*}{\textbf{HOTPOT-QA}}                                                       & RefNet                              & 21.17                                                                                        & 28.7                                                                            \\ \cline{2-4} 
                                                                                 & Reward-RefNet                       & 21.32                                                                                        & 29.2                                                                            \\ \hline
\multirow{2}{*}{\textbf{DROP}}                                                            & RefNet                              & 21.23                                                                                        & 33.6                                                                            \\ \cline{2-4} 
                                                                                 & Reward-RefNet                       & 21.60                                                                                        & 34.3                                                                            \\ \hline
\end{tabular}}
\caption{Impact of Reward-RefNet on various datasets when fluency and answerability are used as reward signals.}
\label{tab:app}
\end{table}

\section{Visualization of Attention Weights}
\begin{figure}[t]
\begin{subfigure}[b]{0.55\textwidth}
   \includegraphics[width=0.9\columnwidth]{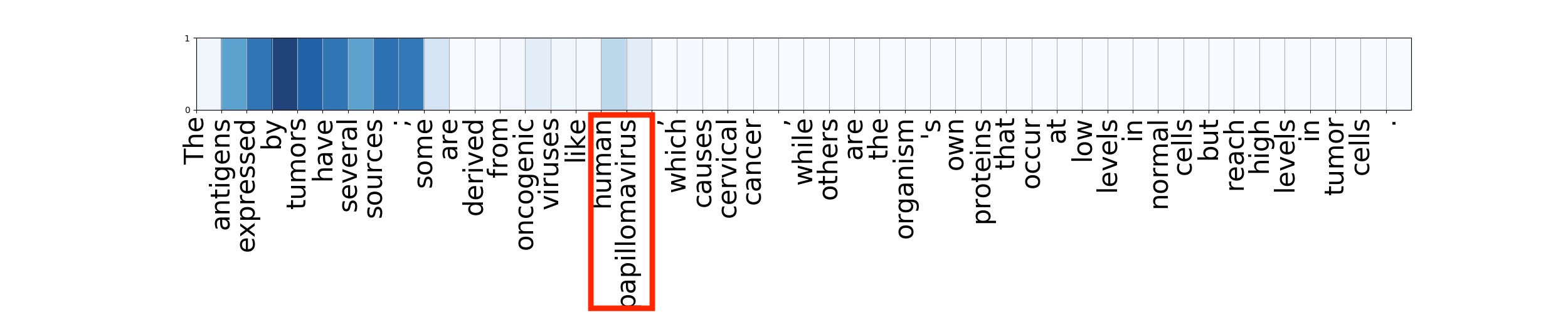}
   \caption{$\mathbf{A_1}$ attention plot}
   \label{fig:Ng1} 
\end{subfigure}
\begin{subfigure}[b]{0.55\textwidth}
   \includegraphics[width=0.9\columnwidth]{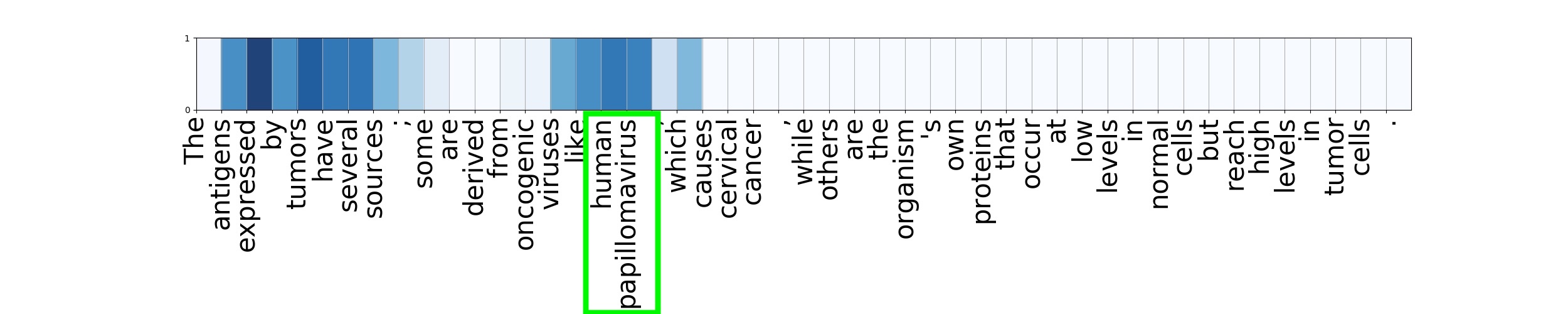}
    \caption{$\mathbf{A_2}$ attention plot}
   \label{fig:Ng2}
\end{subfigure}
\begin{subfigure}[b]{0.55\textwidth}
   \includegraphics[width=0.9\columnwidth]{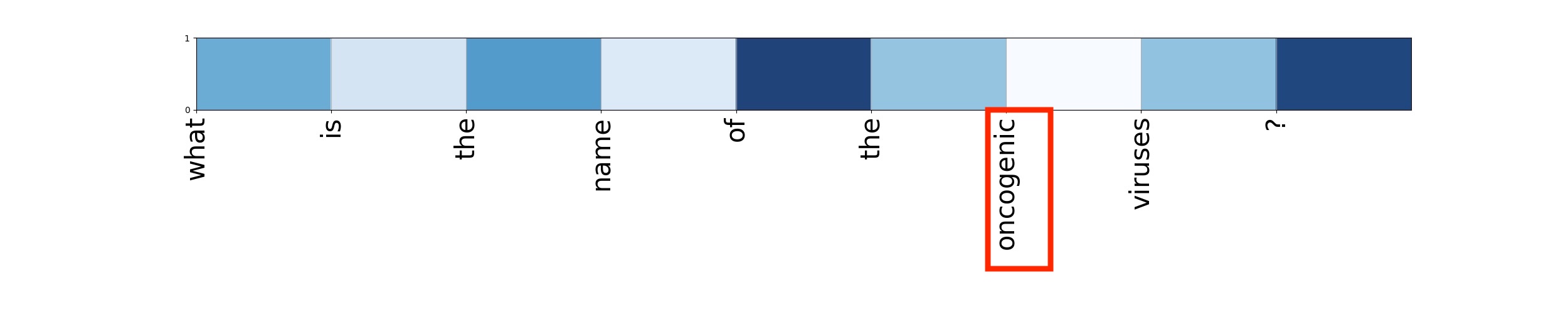}
   \caption{$\mathbf{A_3}$ attention plot}
   \label{fig:Ng2}
\end{subfigure}
\caption{Attention plots for a) $\mathbf{A_1}$, b) $\mathbf{A_2}$, c) $\mathbf{A_3}$ respectively \\
\textbf{Initial Generated Question}: ``What is the name of the oncogenic virus?'' \\
\textbf{Refined Generated Question}: ``What is the name of the organism that causes cervical\\ cancer?''\\
\textbf{Passage}: ``The antigens expressed by tumors have several sources ; some are derived from oncogenic viruses like \textcolor{blue}{human papillomavirus} , which causes cervical cancer , while others are the organism’s own proteins that occur at low levels in normal cells but reach high levels in tumor cells.''}

\label{attentionplots}
\end{figure}

We plot the aggregated attention given to the passage and initial draft of the generation question across the various time-steps of the decoder in Figure \ref{attentionplots}. Although, both the questions are specific to the answer, $\mathbf{A_2}$ pays some attention to the context surrounding the answer, which leads to a complete question. Also, note that in $\mathbf{A_3}$, while attending on to initial draft ``oncogenic'' word is not paid attention to and thus the final draft revises over the initial draft by correcting it to generate a better question.

\end{document}